\documentclass{article} % For LaTeX2e
\usepackage{iclr2026_conference,times}

\usepackage{amsmath,amsfonts,bm}

\def\eqref#1{equation~\ref{#1}}
\def\1{\bm{1}}

\DeclareMathAlphabet{\mathsfit}{\encodingdefault}{\sfdefault}{m}{sl}
\SetMathAlphabet{\mathsfit}{bold}{\encodingdefault}{\sfdefault}{bx}{n}

\usepackage{hyperref}
\usepackage{url}
\usepackage{multirow}
\usepackage{graphicx}
\usepackage{booktabs}
\usepackage{bbding}
\usepackage{xcolor}  
\usepackage{adjustbox}
\usepackage{colortbl}
\usepackage{subfig}
\usepackage{url}
\usepackage{subfloat}
\usepackage{subcaption}
\usepackage[utf8]{inputenc}

\definecolor{mygray}{gray}{0.9}

\title{GranViT: A Fine-Grained Vision Model With Autoregressive Perception For MLLMs}

\author{
  Guanghao Zheng\textsuperscript{1$^{\dagger}$},
  Bowen Shi\textsuperscript{1$^{\dagger}$},
  Mingxing Xu\textsuperscript{2},
  Ruoyu Sun\textsuperscript{2},
  Peisen Zhao\textsuperscript{2},\\
  \textbf{Zhibo Zhang}\textsuperscript{2},
  \textbf{Wenrui Dai}\textsuperscript{1$\ast$},
  \textbf{Junni Zou}\textsuperscript{1},
  \textbf{Hongkai Xiong}\textsuperscript{1}, 
  \textbf{Xiaopeng Zhang}\textsuperscript{2}\thanks{Correspondence to Wenrui Dai and Xiaopeng Zhang. $^{\dagger}$ Equal contribution.},
  \textbf{Qi Tian}\textsuperscript{2} \\
  \textsuperscript{1}Shanghai Jiao Tong University, Shanghai, China \\
  \textsuperscript{2}Huawei Inc., China
}

\iclrfinalcopy % Uncomment for camera-ready version, but NOT for submission.
\begin{document}

\maketitle

\begin{abstract}
Vision encoders are indispensable for allowing impressive performance of Multimodal Large Language Models (MLLMs) in vision–language tasks such as visual question answering and reasoning. However, existing vision encoders focus on global image representations but overlook fine-grained regional analysis. They are limited in fine-grained perception due to the scarcity of fine-grained annotated data and the lack of a fine-grained pre-training paradigm. In this paper, we propose GranViT, a novel Vision Transformer that integrates fine-grained feature extraction with semantic alignment to Large Language Models (LLMs) via region-level autoregressive training. We first construct \emph{Gran-29M}, a dataset comprising 29 million natural and OCR images paired with over 180 million high-quality region-level annotations, to enable large-scale fine-grained pretraining. Consequently, we develop a pretraining-adaptation framework along with a self-distillation mechanism to train fine-grained \emph{GranViT} on \emph{Gran-29M}. We sufficiently exploit the fine-grained annotations from \emph{Gran-29M} to resort to bounding-box-to-caption regression to enhance localized visual representation of the vision encoder in the pretraining and caption-to-bounding-box regression to improve vision feature utilization and localization for LLM in the adaptation. 
%\emph{Gran-29M} enables complementary pretraining tasks for tuning vision encoder and LLM, \emph{i.e.}. bounding-box-to-caption regression to enhance localized visual representation of vision encoder and caption-to-bounding-box regression to improve LLM localization. 
We further incorporate a self-distillation mechanism that imposes explicit localization constraints on the vision encoder to strengthen its regional reasoning capability. Extensive experiments show that GranViT surpasses existing vision encoders and attains strong transferability to varying LLMs. Remarkably, it achieves state-of-the-art results on fine-grained recognition, multimodal VQA, and OCR understanding. %We believe GranViT represents an important advancement in vision foundation models. All codes and pretrained weights will be released upon acceptance.

% New Version
% We present GranViT, a vision foundation model designed to advance fine-grained general image and OCR understanding. W

\end{abstract}

\section{Introduction}
Multimodal Large Language Models (MLLMs) have stimulated substantially growing research interests and efforts in recent years \citep{wang2024qwen2,bai2025qwen2,dong2025scalable,zhu2025internvl3,wang2025internvl3,li2025eagle}. Existing architectures for MLLMs usually consist of a pretrained vision encoder that extracts visual information and a projection module that maps visual information to visual tokens for image understanding and reasoning with Large Language Models (LLMs). %The adapter module (e.g., a projector \citep{liu2023visual} or Q-Former \citep{li2023blip}) serves as a bridge that maps visual features into the semantic space of the LLMs, thereby achieving cross-modal feature alignment. 
Projection modules such as multilayer perceptrons (MLPs) \citep{liu2023visual} or Q-Formers \citep{li2023blip}  bridge visual features to the semantic space of LLMs, whereas vision encoders are primarily for the ability of capturing visual information for MLLMs.

Vision Transformers (ViTs) \citep{dosovitskiy2020image} and their variants \citep{liu2021swin,ravi2024sam} have been widely adopted as vision encoders in MLLMs due to their exceptional capabilities in visual feature extraction and scalability \citep{dosovitskiy2020image,carion2020end, kirillov2023segment,ravi2024sam}.
% , as evidenced by their strong performance in traditional vision tasks
%such as classification \citep{dosovitskiy2020image}, object detection \citep{carion2020end,dosovitskiy2020image}, and segmentation \citep{kirillov2023segment,ravi2024sam}.
Existing ViTs are usually trained to align visual representations with textual semantics. Contrastive Language–Image Pre-training (CLIP) \citep{radford2021learning,zhai2023sigmoid,tschannen2025siglip,shi2024umg} is one prevailing paradigm that projects images and texts into a learned shared embedding space to aggregate matched image-text pairs, and non-matching pairs are discriminated to preserve the semantic relationship.
Another popular alternative is autoregressive modeling \citep{chen2024internvl,fini2025multimodal,tschannen2025siglip} that directly maps visual features into the textual space by connecting a cascaded vision encoder and textual decoder. It allows superior alignment with textual space but could sacrifice the discrimination ability of visual features.
% Furthermore, some approaches integrate both methods \citep{tschannen2025siglip,chen2024internvl}, jointly training contrastive learning and autoregressive modeling to simultaneously preserve the discriminative nature of visual features and their alignment with the textual space.
Nevertheless, these approaches over-emphasize image-level global feature extraction but neglect essential fine-grained details required for multimodal understanding. 

To address the limitation, in this paper, we investigate integrating fine-grained localization capabilities into the vision encoder within an LLM cascade architecture. It is non-trivial to address the following two challenges, \emph{i.e.}, i) Data scarcity: scarcity of high-quality datasets with fine-grained annotations, and ii) Fine-grained pre-training: lack of a dedicated framework to train fine-grained vision encoders that effectively align with LLMs. 

\textbf{i) Data scarcity.} We build a high-quality annotated dataset termed \emph{Gran-29M} that contains 29 million natural and OCR images with image-level annotations along with 183 million region-level annotations. Specifically, we leverage the UMG-41M \citep{shi2024umg}, FLICKR30k \citep{young2014image}, and LAION \citep{schuhmann2022laion} datasets to collect natural images of varying scales and diversity and generate image-level and region-level annotations (e.g., bounding boxes) using ViTDet \citep{li2022exploring} and Qwen2.5-VL \citep{bai2025qwen2}. Moreover, we consolidate publicly available OCR datasets \citep{li2025eagle,li2024llavaonevision} and utilize PaddleOCR \citep{cui2025paddleocr} for localized text detection and bounding box prediction. \emph{Gran-29M} is achieved with rigorous filtering based on bounding box aspect ratio, area, quantity, and image resolution.

%To meet the data demands, we developed a high-quality dataset termed Gran-29M, which contains both global and localized annotations for natural and OCR images. For natural images, we leverage the UMG-41M \citep{shi2024umg}, FLICKR30k \citep{young2014image}, and LAION \citep{schuhmann2022laion} datasets to ensure scale and diversity, while generating image-level and region-level annotations (e.g., bounding boxes) using ViTDet \citep{li2022exploring} and Qwen2.5-VL \citep{bai2025qwen2}. For OCR images, we consolidate publicly available OCR datasets \citep{li2025eagle,li2024llavaonevision} and utilize PaddleOCR \citep{cui2025paddleocr} for localized text detection and bounding box prediction. Through rigorous filtering based on bounding box aspect ratio, area, quantity, and image resolution, we curate an annotated dataset of 29 million images with 183 million region-level annotations.

\begin{figure}[!t]
    \centering
    \subfloat[]{
    \includegraphics[height=6.5cm,width=7cm]{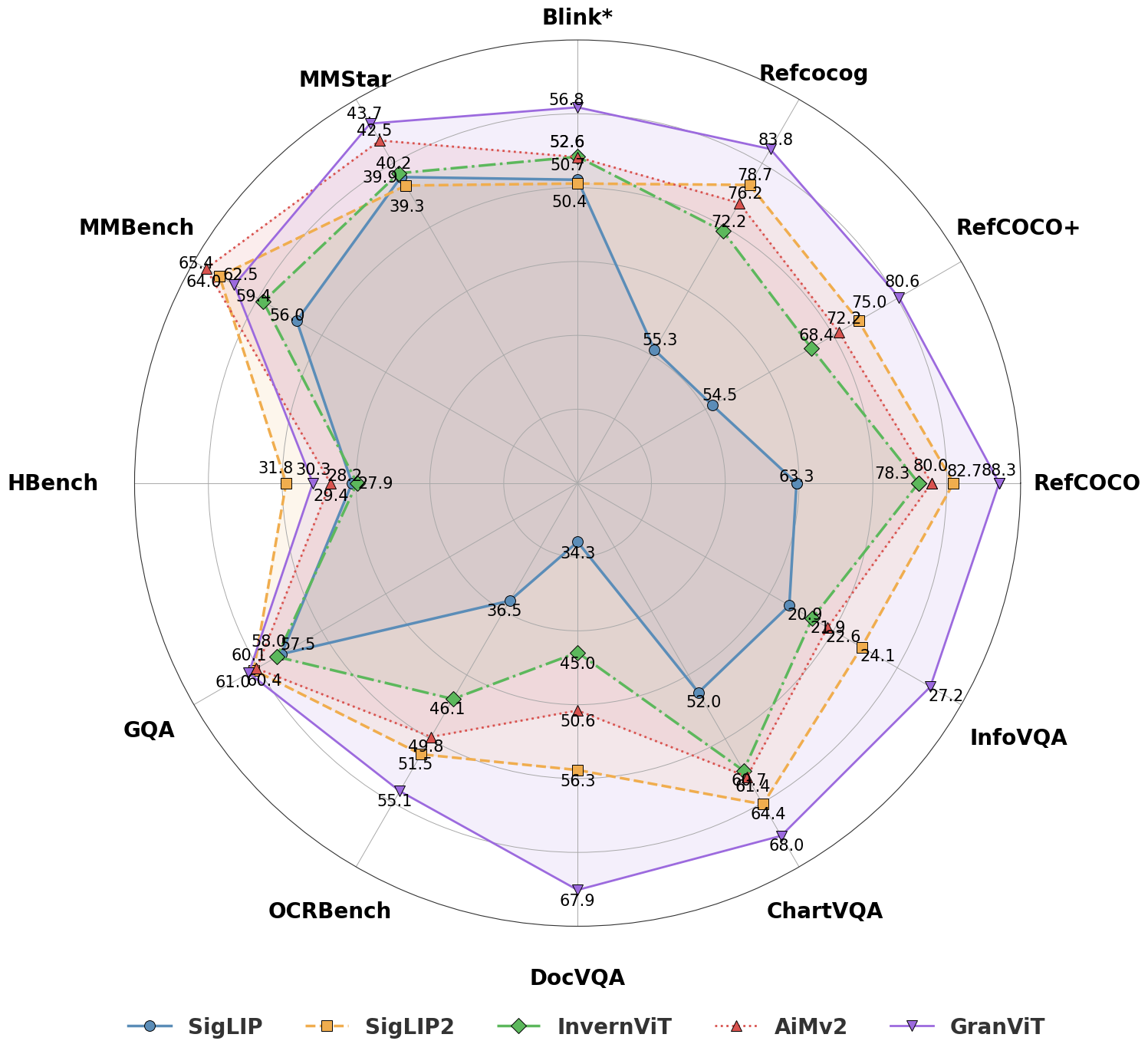}
    }
    \subfloat[]{
    \includegraphics[height=6cm,width=6cm]{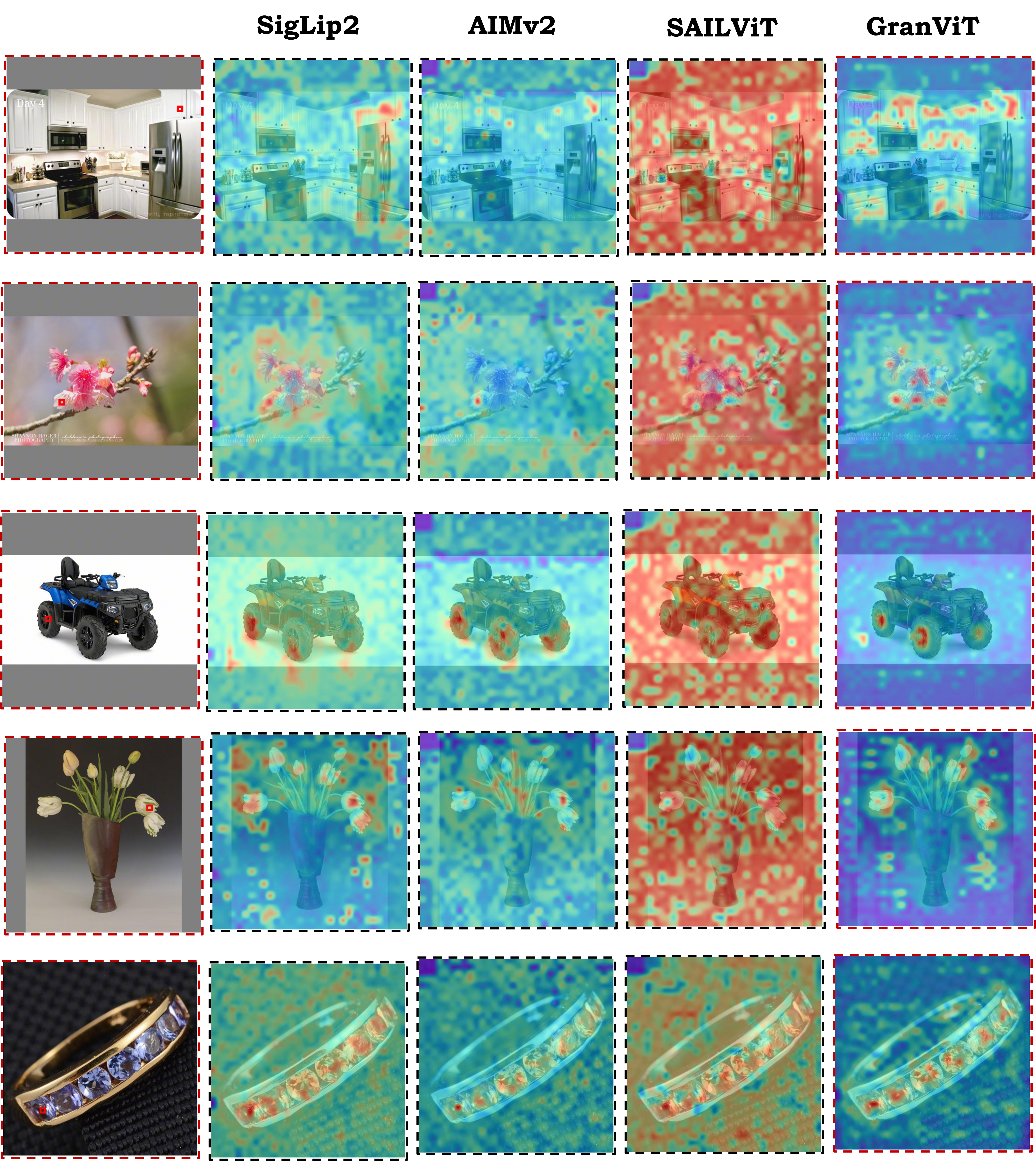}
    }
\vspace{-0.2cm}
    \caption{(a) Compared to existing vision encoders, GranViT demonstrates outstanding performance across fine-grained natural image and OCR understanding. HBench denotes HallusionBench. (b) Attention visualization of existing vision encoders according to the query token. The small red rectangle indicates the query token. Best viewed with zoom in.}
    \label{fig:1}
    \vspace{-0.8cm}
\end{figure}

% \begin{figure}[!t]
%     \centering

% \vspace{-0.3cm}
%     \caption{Attention visualization of existing vision encoders according to the query token. The small red rectangle indicates the query token. Best viewed in zoom in.}
%     \label{fig:5}
% \end{figure}

\textbf{ii) Fine-grained pre-training and adaptation.} We propose a novel pretraining-adaptation framework to improve fine-grained understanding of natural and OCR images beyond enhancing overall perception. In the pretraining, the proposed framework optimizes the vision encoder with bounding-box-to-caption ($Bbox2Caption$) regression for fine-grained feature extraction. Additionally, we develop localized self-distillation to optimize the vision encoder and explicitly augment its ability to extract fine-grained features. As for adaptation, the LLM is tunable for fine-grained vision feature localization with caption-to-bounding-box ($Caption2Bbox$) regression.

%Moreover, beyond enhancing overall perception of both natural and OCR images, we introduce a novel pretraining framework to improve fine-grained understanding of these data types. Implemented in two stages, the framework performs bounding-box-to-caption ($Bbox2Caption$) and caption-to-bounding-box ($Caption2Bbox$) tasks to optimize the vision encoder and the LLM, respectively. Specifically,
% we propose a novel pretraining framework that enhances the fine-grained capabilities of both the vision encoder and the LLM in two distinct stages with $Bbox2Caption$ and $Caption2Bbox$ tasks, respectively. 
%in the first stage, the vision encoder is optimized to produce features that enable the LLM to localize and recognize regional visual elements accurately. During the second stage, the LLM is trained to leverage these visual features for object recognition and localization. Furthermore, the localized self-distillation is additionally introduced in the first stage to explicitly augment the fine-grained feature extraction capability. 

To validate the effectiveness of GranViT, we perform comprehensive performance comparisons and extensive visualizations after downstream supervised fine-tuning (SFT) \citep{li2024llava}, including visual question answering, visual grounding, and OCR understanding. Fig.~\ref{fig:1} shows that GranViT achieves state-of-the-art performance on multiple benchmarks and exhibits strong generalization capabilities. The contributions of this work are summarized as below.

$\bullet$ We establish \emph{Gran-29M}, a large-scale pretraining dataset containing 29 million natural and OCR images with comprehensive global annotations and 183 million fine-grained captions.

$\bullet$ We propose a pretraining-adaptation framework that simultaneously enhances the fine-grained feature extraction ability of GranViT with $Bbox2Caption$ regression and localized self-distillation using explicit local region supervision and adapts to varying LLMs with stronger capacity for local region localization with $Caption2Bbox$ regression. %Besides, localized self-distillation is applied in 1st stage for explicit local region supervision. Through this framework, we develop GranViT, a novel vision encoder exhibiting enhanced fine-grained feature extraction capacity and strong transferability.

$\bullet$ We demonstrate the robustness and generalization ability of \emph{GranViT} compared with existing vision encoders via comprehensive analysis.  \emph{GranViT} achieves state-of-the-art performance on visual grounding and OCR comprehension.
% We demonstrate the robustness and generalizability of GranViT through a comprehensive analysis, covering various parameter sizes, model architectures, training strategies, and data scales. 
%Compared to other vision encoders, GranViT exhibits enhanced capabilities in visual question answering, visual grounding, and OCR comprehension.

\vspace{-0.3cm}
\section{Related Work}
\vspace{-0.2cm}
\subsection{Multimodal Large Language Models}
\vspace{-0.2cm}
% Building on the powerful textual understanding and reasoning capabilities of large language models (LLMs) \citep{touvron2023llama,yang2025qwen3,guo2025deepseek}, 
Multimodal large language models (MLLMs) \citep{wang2024qwen2,bai2025qwen2,dong2025scalable,chen2024internvl,chen2024expanding,zhu2025internvl3,wang2025internvl3,team2025kwai,li2025eagle,lei2025medlsam} attract wide attention for their potential in image understanding and reasoning.
Building on the robust textual understanding and reasoning capabilities of large language models (LLMs) \citep{touvron2023llama,yang2025qwen3,guo2025deepseek,yang2025qwen3,lei2025data}, most existing MLLMs augment their functionality with a pretrained vision encoder to enable visual perception. These encoders are usually trained with contrastive learning \citep{radford2021learning,zhai2023sigmoid,tschannen2025siglip,shi2024umg} and projectors commonly adopt a two-layer MLP architecture \citep{li2024llava,liu2023visual}, but pre-trained vision encoders cannot handle high-resolution inputs \citep{zhai2023sigmoid}.
%Due to their strong semantic modeling abilities, many MLLMs adopt the Qwen series \citep{yang2025qwen3,team2024qwen2,bai2023qwen} as the language backbone.
% Early models such as Qwen2-VL \citep{wang2024qwen2}, Aquila-VL \citep{gu2024infinity}, DeepSeek-VL \citep{lu2024deepseek}, and Keye-VL \citep{team2025kwai} employ a pre-trained SigLIP \citep{zhai2023sigmoid} vision encoder and conduct direct image-text alignment with supervised fine-tuning (SFT). 
%Pre-trained vision encoders cannot handle high-resolution inputs \citep{zhai2023sigmoid}. 
Early models \citep{wang2024qwen2,gu2024infinity} adopt an image tiling strategy \citep{chen2024internvl,liu2023visual,lu2024deepseek,team2025kwai}: high-resolution images are divided into patches, from which local features are extracted and aggregated.
In comparison, newer MLLMs such as Qwen2.5-VL \citep{bai2025qwen2}, Seed-VL1.5 \citep{guo2025seed1}, and Kimi-VL \citep{team2025kimi} train vision encoders from scratch on diverse datasets and support native-resolution input \citep{bai2025qwen2} to mitigate performance loss from resolution reduction.
Beyond architectural improvements, recent MLLMs increasingly focus on post-training strategies \citep{cheng2024domain,gu2024infinity,li2025eagle}. These emphasize large-scale, curated SFT datasets and leverage both SFT and reinforcement learning \citep{schulman2017proximal,shao2024deepseekmath,zheng2024beta,zheng2025bidirectional,zheng2024mc} to enhance task-specific capability.

\vspace{-0.2cm}
\subsection{Vision Foundation Models}
\vspace{-0.2cm}
Vision encoders are a critical component for extracting and representing visual information to support multimodal reasoning in MLLMs. Existing MLLMs usually employ vision encoders pre-trained through contrastive learning, which inherently align visual and textual semantic spaces. Commonly used encoders include CLIP \citep{radford2021learning} using cross-entropy loss \citep{mao2023cross}, and SigLIP \citep{zhai2023sigmoid} using sigmoid loss. InternViT \citep{chen2024internvl} combines contrastive learning with an autoregressive loss and incorporates a text decoder to enhance alignment by decoding visual features into text. SeedViT \citep{guo2025seed1} first undergoes generative self-supervised pretraining \citep{xie2022simmim,he2022masked} before contrastive learning. AIMv2 \citep{fini2025multimodal} introduces the first vision encoder trained solely with an autoregressive loss, predicting subsequent image patches and text tokens to achieve cross-modal alignment without contrastive learning. %, thereby improving contextual modeling and generalization. 
SigLIP2 \citep{tschannen2025siglip} integrates autoregressive and self-distillation losses in SigLIP to enhance visual representations through multi-objective pretraining. SAILViT \citep{yin2025sailvit} extends AIMv2 by incorporating alignment with LLMs and multi-stage pretraining with SFT data, facilitating high-dimensional vision-language integration and infusing world knowledge into visual encoding. However, these encoders predominantly emphasize global feature extraction at the cost of fine-grained visual details, and are limited in fine-grained multimodal tasks.

\section{Gran-29M: Fine-grained Annotated Dataset}
In this section, we elaborate on the construction of a large-scale fine-grained \emph{Gran-29M} dataset for pre-training, including data sources, data annotations, filtering criteria, and data reformatting.

\begin{figure}[!t]
    \centering
\includegraphics[height=3cm,width=13.5cm]{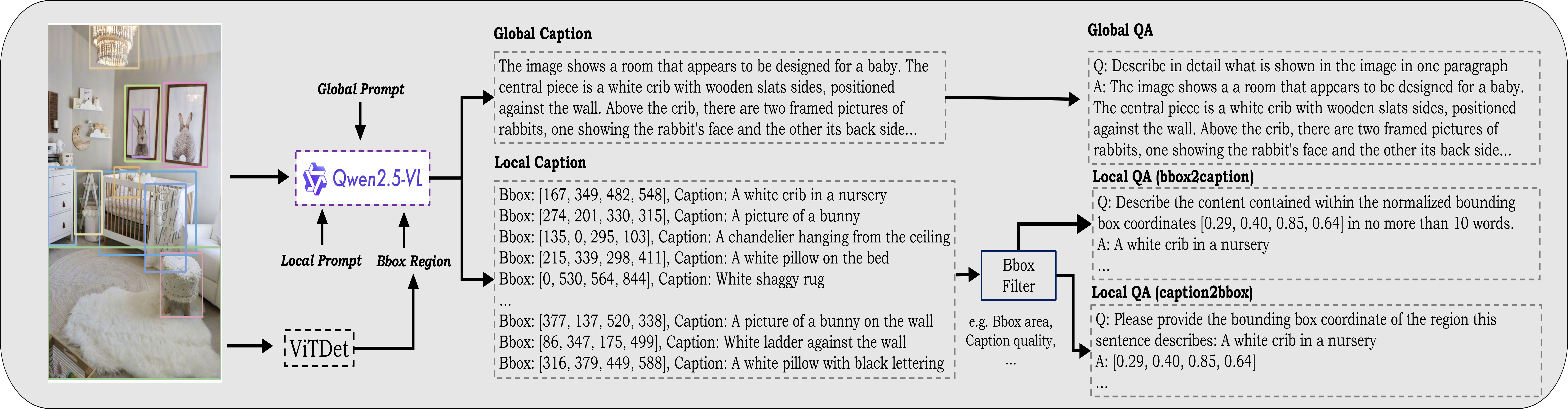}
\vspace{-0.3cm}
    \caption{The details of data annotations of \emph{Gran-29M}. We leverage ViTDet \citep{li2022exploring} and Qwen2.5-VL-7B \citep{bai2025qwen2} for bbox and caption generation. Then, we transfer the absolute bbox coordinate to a relative one based on the image resolution and apply rigorous filtering based on image resolution, bbox area, and the number of bboxes per image. Finally, we reformat the global and local captions into QA pairs. }
    \label{fig:2}
\vspace{-0.3cm}
\end{figure}

% \begin{table}[!t]
% \renewcommand{\baselinestretch}{1.0}
% \renewcommand{\arraystretch}{1.2}
% \setlength{\tabcolsep}{8.0pt}
% \setlength{\abovecaptionskip}{15pt}
% \centering
% \caption{Data sources of natural and OCR images in Gran-29M. $\#images$ and $\#regions$ denote the number of images and annotated bounding boxes after filtering, respectively.}\label{table1}
% \begin{tabular}{c l r r}
% \hline
% Data Type & Data Source & $\#images$ & $\#regions$ \\
% \hline
% % \multirow{8}{*}{Natural} & CC3M  & 565K & 2.3M \\
% %                          & IN21k & 614K & 1.6M \\
% %                          & LAION & 17.19M & 54.35M \\
% %                          & SBU & 21K & 52K \\
% %                          & CC12M  & 4.9M & 21.74M \\
% %                          & FLICKR30k  & 1269 & 4351 \\
% %                          & YFCC15M & 655K & 1.8M \\
% %                          & VisualGenome  & 2150 & 14825 \\

% \multirow{3}{*}{Natural} & UMG-41M  & 6.7M & 27.63M \\
%                          & LAION & 17.19M & 54.35M \\
%                          & FLICKR30k  & 1269 & 4351 \\
                         
% \hline
% \multirow{4}{*}{OCR}  & Text Images & 1.9M & 42.57M \\
%                       & Chart, Table & 325K & 2.8M \\
%                       & Invoice Receipt & 2847 & 41K \\
%                       & Rich Text Images & 3.3M & 56M \\
% \hline
% & TOTAL & 29.51M & 183.55M \\
% \hline
% \end{tabular}
% \vspace{-0.5cm}
% \end{table}

\textbf{Data Source.} 
%As shown in Table~\ref{table1}, 
We collect diverse large-scale images from public datasets. For natural images, we expand the UMG-41M dataset \citep{shi2024umg} (including CC3M \citep{sharma2018conceptual}, IN21k \citep{deng2009imagenet}, SBU \citep{ordonez2011im2text}, CC12M \citep{changpinyo2021conceptual}, YFCC15M \citep{kamath2021mdetr}, and VisualGenome \citep{krishna2017visual}) with samples from LAION \citep{schuhmann2022laion} and FLICKR30k \citep{young2014image}. For OCR images, we collect four distinct types of images from publicly available sources \citep{li2024llavaonevision, li2025eagle}: 
% For OCR data, we gathered four major categories from various SFT datasets \citep{li2024llavaonevision,li2025eagle}: 
plain text images, chart and table images, receipt images, and rich text images. Refer to Table~\ref{table-a1} in the appendix for details. 

\textbf{Data Annotations Workflow.} For natural images, we directly utilize bounding box (bbox) coordinates provided by UMG-41M as localized annotations for local regions, and employ Qwen2.5-VL-7B \citep{bai2025qwen2} to regenerate global and local captions to enhance caption quality. For the LAION dataset \citep{schuhmann2022laion} and FLICKR30k \citep{young2014image}, we utilize ViTDet \citep{li2022exploring} to detect bboxes and Qwen2.5-VL-7B \citep{bai2025qwen2} to generate global and local captions, as shown in Fig.~\ref{fig:2}. For OCR images, since global descriptions are often vague (\emph{e.g.}, ``a page of an academic paper'') and lack details, only local regions are annotated using PaddleOCR \citep{cui2025paddleocr} to provide accurate bboxes and textual contents simultaneously.

\textbf{Filtering Criteria.} To ensure high-quality annotations for both global and local regions, we apply a filtering process based on image resolution and bbox criteria. For local region annotations, we require that the shorter side of each image should be larger than $448$ pixels, the aspect ratio of both the entire image and each bbox should be between $\frac{1}{3}$ and $3$, the area of each bbox should be greater than $100^2$ square pixels, and the number of bboxes per image should be at least one. The filtered results are summarized in Table~\ref{table-a7} in the appendix. In total, we obtain 29.51 million images with 183.55 million localized region annotations for large-scale pretraining.

\textbf{Data Reformatting.} To facilitate the training of GranViT, we reorganize existing global and local region captions and reformat them into a standard question-answer pair structure. Using the following question and answer prompts, we rewrite existing data to enhance its suitability for training. For bbox coordinates and corresponding captions, we perform bidirectional annotations through $Bbox2Caption$ and $Caption2Bbox$ tasks for the vision encoder and LLM pretraining, respectively. Furthermore, we convert the absolute bbox coordinates into relative coordinates based on image resolutions to eliminate the dependence on absolute coordinates.
\begin{description}
\vspace{-0.2cm}
\item[Global Caption.] \hfill\\
        \begin{minipage}{0.85\textwidth}
            \texttt{Q: Describe in detail what is shown in the image in one paragraph} \\
            \texttt{A: [global captions]}
        \end{minipage}
\item[Bbox2Caption.] \hfill\\
\begin{minipage}{0.85\textwidth}
\texttt{Q: Describe the content contained within the normalized bounding box coordinates [bbox coordinates] in no more than 10 words.} \\
\texttt{A: [local captions]}
\end{minipage}   
\item[Caption2Bbox.] \hfill \\
\begin{minipage}{0.85\textwidth}
\texttt{Q: Please provide the bounding box coordinate of the region this sentence describes: [local captions]} \\
            \texttt{A: [bbox coordinations]}
        \end{minipage}
\vspace{-0.2cm}
\end{description}

\begin{figure}[!t]
    \centering
\includegraphics[height=7cm,width=13.5cm]{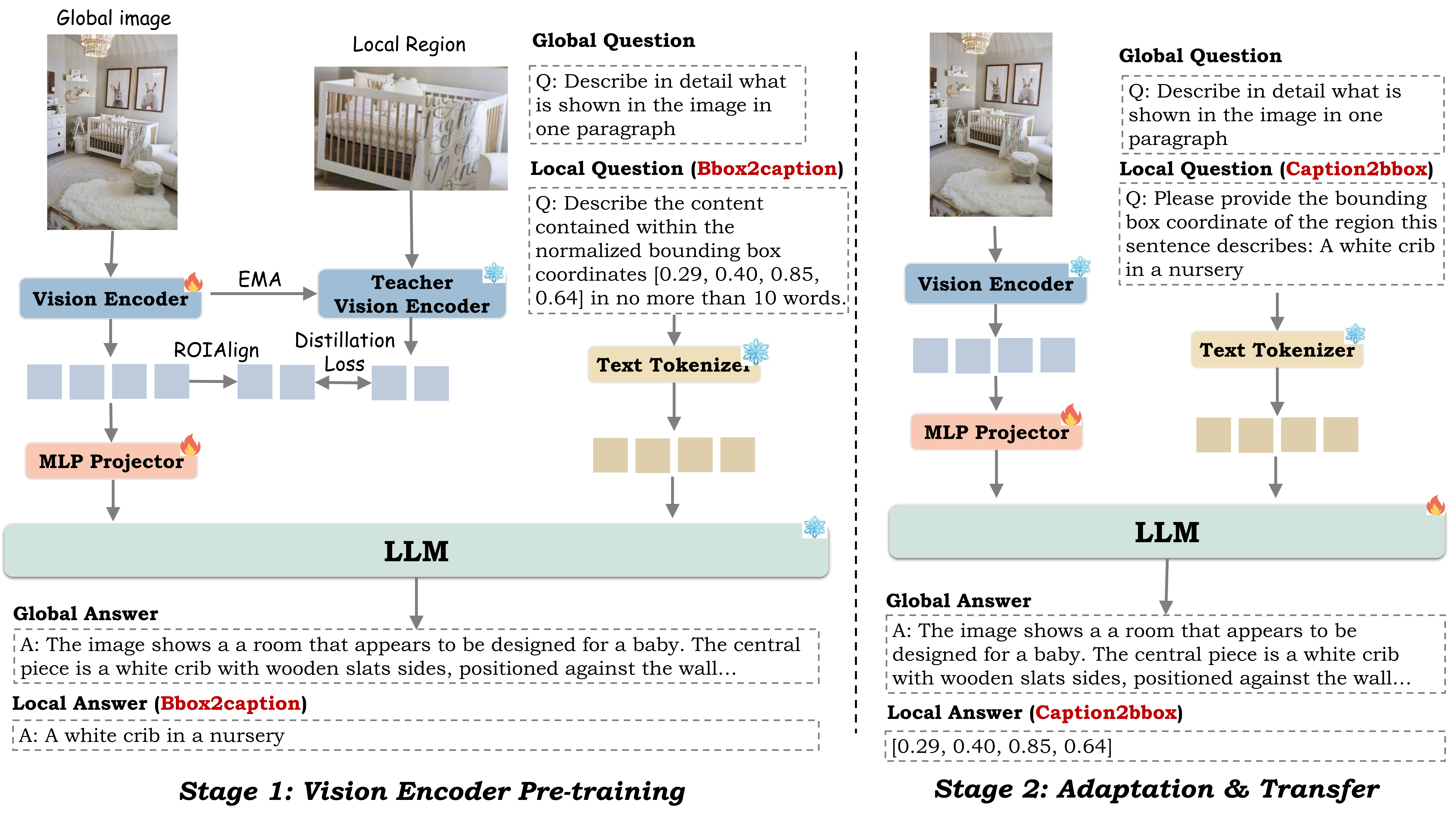}
\vspace{-0.3cm}
    \caption{The fine-grained pretraining and transferring paradigm of GranViT. For pretraining, the vision encoder and projector are tuned via the global and $Bbox2Caption$ task for fine-grained feature extraction. The teacher vision encoder explicitly supervises the local region of features extracted by the student vision encoder. For vision feature adaptation and transfer, based on the fine-grained vision encoder, we apply LLM tuning to further strengthen the localization capability of the LLM regarding fine-grained visual features via the global and $Caption2Bbox$ task.}
    \label{fig:3}
\vspace{-0.5cm}
\end{figure}

\section{Proposed Method}
% In this section, we present the fine-grained pretraining paradigm with autoregressive perception for vision encoders. The approach first trains the vision encoder to improve its fine-grained feature extraction ability. It then trains the LLM to better leverage these features and enhance its local region localization performance. Moreover, self-distillation is employed to further improve the fine-grained representation learning of the vision encoder.

\subsection{Fine-Grained Pretraining Paradigm with Autoregressive Perception}
Owing to training solely on images and global captions \citep{radford2021learning,zhai2023sigmoid,fini2025multimodal}, previous vision encoders struggle with fine-grained feature extraction for local regions, while also lacking alignment between visual features and the textual feature of the LLM. 
To overcome these issues with a unified framework, we leverage the LLM to provide supervision for the fine-grained training of vision encoders. Specifically, we employ the same global image captioning task \citep{radford2021learning,zhai2023sigmoid,fini2025multimodal} throughout the entire pretraining process to preserve the global perception capability. Furthermore, we enhance the ability of fine-grained feature extraction by cascading the vision encoder with the LLM via the projector during pretraining, and perform large-scale pretraining using both $Bbox2Caption$ and $Caption2Bbox$ tasks for localized region recognition and grounding, respectively.  As depicted in Fig.~\ref{fig:3}, the proposed framework consists of pretraining and adaptation stages.
\begin{itemize}
\item \textbf{Stage 1: Pretraining that tunes vision encoder and projector with LLM frozen.} We additionally employ the $Bbox2Caption$ task for pretraining, which requires the MLLM to generate a localized caption of the object within specified bboxes. The LLM can be viewed as a decoder that converts visual features into texts, where the supervision is directly propagated back to the extracted local features with bboxes, thereby enhancing the fine-grained characteristics of the visual representations. The input prompt to the LLM incorporates the bbox coordinates, thereby facilitating object recognition and localization. It is worth noting that, in this stage, we employ a lightweight LLM (\emph{i.e.}, Qwen2.5-VL-1.5B \citep{bai2025qwen2}) to compel the vision encoder to extract generic fine-grained features, rather than relying on the powerful reasoning capabilities of large LLMs for output generation.
\item \textbf{Stage 2: Adaptation and Transfer that tune projector and LLM with the vision encoder frozen.} In contrast, we employ the $Caption2Bbox$ task in this stage, which requires the MLLM to recognize objects present in the image according to the prompts and output their bbox coordinates. The primary objectives of this stage are to further enhance the localization capability of the LLM based on fine-grained visual features and to ensure the transferability of the vision encoder to other LLMs with comparable or larger size. Since in MLLMs, it is required that the vision encoder provide more fine-grained features, while the LLM should also be capable of utilizing these visual features. On the other hand, the vision encoder can be adapted with different LLMs (\emph{i.e.}, Qwen2.5-VL-3B, Qwen2.5-VL-7B \citep{bai2025qwen2}), ensuring compatibility and transferability to new architectures.
\end{itemize}

Across both stages, the autoregressive caption loss is applied to regulate the text output of the LLM for supervising the vision encoder and the LLM, respectively. Given ground truth text $T$ and output text $O_{LLM}$, the caption loss can be calculated by $L_{caption}=CrossEntropy(O_{LLM},T)$.
%, where $CrossEntropy$ denotes the Cross-Entropy Loss. 
Through this pretraining-adaptation pretraining paradigm, the vision encoder gains enhanced fine-grained feature extraction abilities, with its outputs inherently aligned to the semantic space of the LLM. The LLM, in turn, improves its capacity to utilize visual information for accurate localization. These generalized capabilities enhance object recognition and spatial reasoning during downstream SFT, thereby reducing visual understanding errors and mitigating hallucinations.

% \begin{figure}[!t]
%     \centering
%     \adjustbox{valign=t}{\includegraphics[height=5cm]{iclr2026/framework/stage1_curves.png}}
%     \adjustbox{valign=t}{\includegraphics[height=5cm]{iclr2026/framework/Caption2Bbox_comp.png}}
%     \caption{(a) Training curves for Stage1 and Stage2; (b) Bbox visualization of $Caption2Bbox$ tasks.}
%     \label{fig:3}
% \end{figure}

\begin{figure}[!t]
    \centering
    \subfloat[]{
\includegraphics[height=5cm]{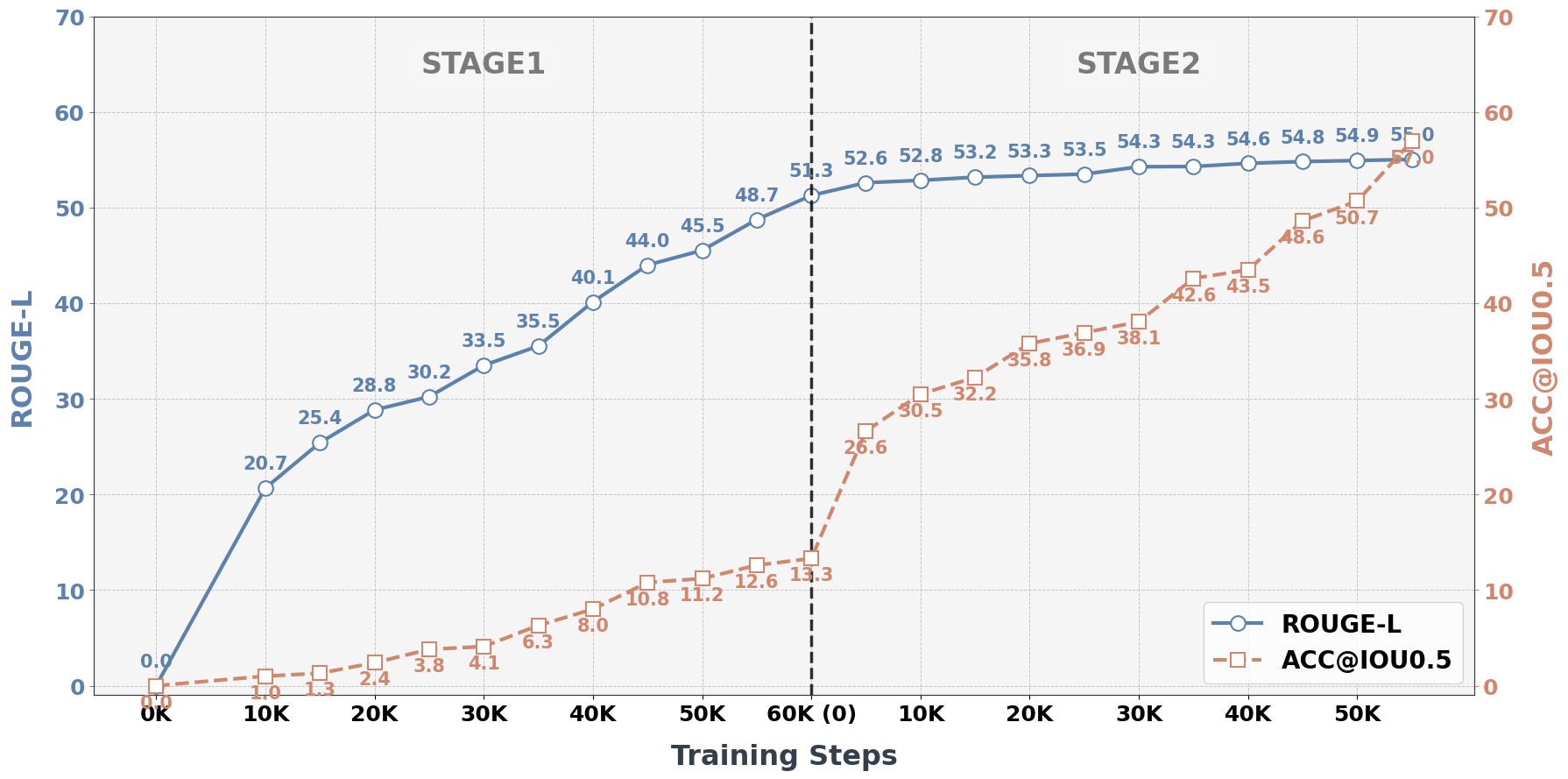}
}
    \subfloat[]{
    \includegraphics[height=5cm]{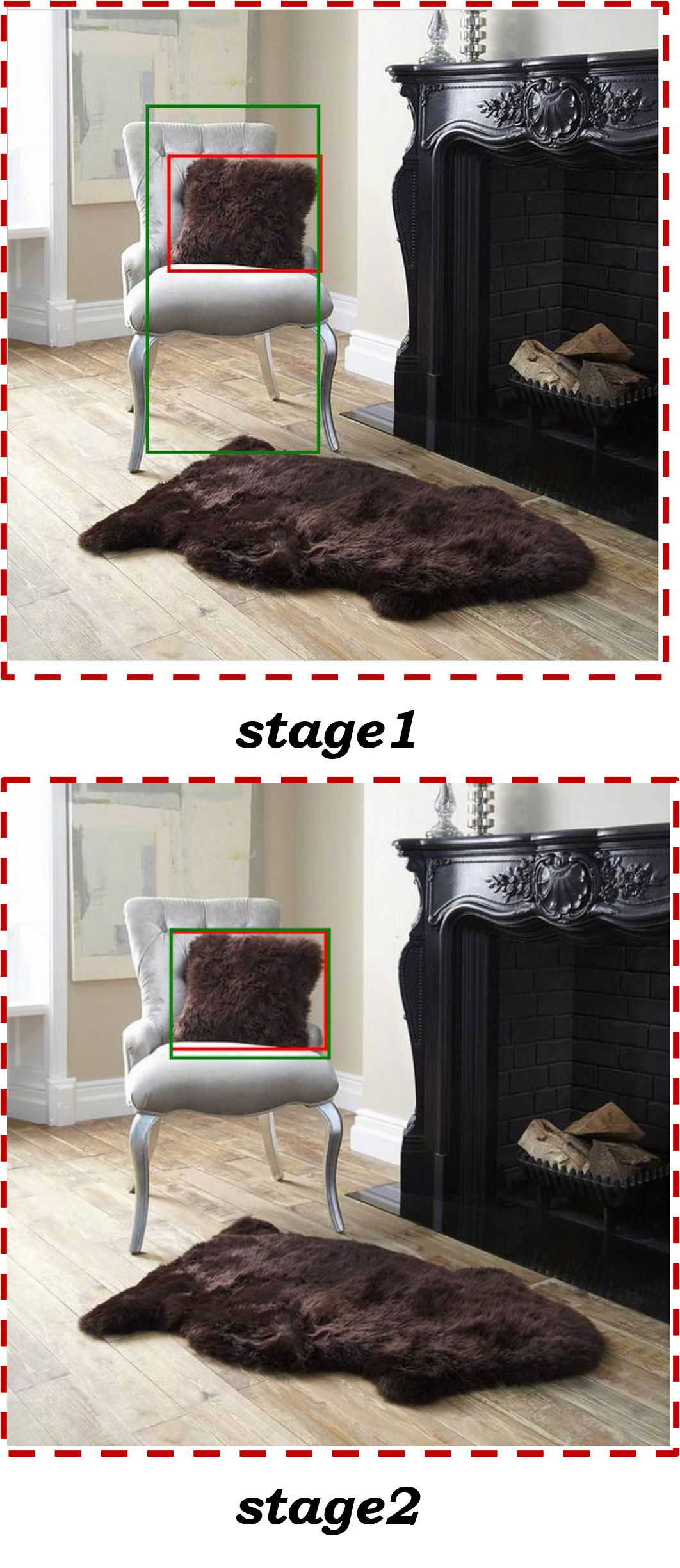}
    }
    \vspace{-0.3cm}
    \caption{(a) The performance curve of Stage1 and Stage2. We sample 8M $Bbox2Caption$ and $Caption2Bbox$ samples respectively for pretraining and adaptation and calculate ROUGE-L \citep{barbella2022rouge} and ACC@IOU0.5 for $Bbox2Caption$ and $Caption2Bbox$ respectively. In stage 1, the ACC@IOU0.5 of the $Caption2Bbox$ task only achieves $13\%$, while the ROUGE-L of the $Bbox2Caption$ task achieves $52\%$. Conversely, in stage 2, the training of LLM leads to a notable increase in ACC@IOU0.5 for $Caption2Bbox$, while $Bbox2Caption$ achieves only a minimal improvement of $3\%$. (b) Visualization of predicted bbox coordinate of $Caption2Bbox$ task in stage 1 and stage 2. Green bboxes indicate predicted regions, while red ones denote ground truth.}
    \label{fig:4}
    \vspace{-0.5cm}
\end{figure}

\subsection{Explanation of Fine-Grained Pretraining Paradigm}
% We provide a detailed explanation of the two-stage paradigm and conduct corresponding ablation studies to validate both the effectiveness and efficiency of the proposed pretraining paradigm.

\textbf{Pretraining-Adaptation framework with different tasks.} 
To enhance the fine-grained feature extraction capability of the vision encoder, in pretraining, we freeze the LLM and only train the vision encoder and projector in stage 1.
% , leveraging text-conditioned supervision to propagate gradients back to the vision encoder and align the visual and textual semantic spaces. 
Both the $Bbox2Caption$ and $Caption2Bbox$ tasks are used initially; however, as illustrated in Fig.~\ref{fig:4}(a), the model shows strong performance in $Bbox2Caption$ but limited accuracy in $Caption2Bbox$, primarily due to the frozen LLM, which restricts learning for the more language-reliant $Caption2Bbox$. Thus, we further apply stage 2 to adapt the pretrained vision encoder to LLMs to utilize fine-grained visual features. Meanwhile, the pretrained vision encoder can be adapted with other LLMs for transferability. Specifically, in stage 2, we keep the vision encoder frozen and tune the projector and LLM using both $Bbox2Caption$ and $Caption2Bbox$ tasks. Results in Fig.~\ref{fig:4}(a) demonstrate a notable improvement in ACC@IOU0.5 for $Caption2Bbox$, while $Bbox2Caption$ sees only marginal gains. This finding is consistent with the visualization results in Fig.~\ref{fig:4}(b), confirming the superiority of training $Caption2Bbox$ in stage 2. Since further optimizing $Bbox2Caption$ in stage 2 yields minimal benefits at twice the computational cost, we finally adopted the pretraining-adaptation paradigm, separately optimizing the two tasks in these two stages.
%the vision encoder is already optimized for $Bbox2Caption$ in stage 1, further joint training 

% We attribute this performance gap to inherent differences between the two localized training tasks. In Bbox2Caption, the LLM can already extract visual features when given a bounding box; the main challenge, therefore, lies in enabling the vision encoder to produce sufficiently fine-grained features. In contrast, for the Caption2Bbox task, the LLM lacks inherent localization capabilities prior to multimodal training. Consequently, with the LLM frozen in the first stage, $Caption2Bbox$ performance remains constrained. Upon unfreezing the LLM in stage 2, however, its ability to localize based on text is effectively enhanced, leading to substantial gains on the task.
% Meanwhile, since the vision encoder was already optimized in stage 1 to generate LLM-compatible fine-grained features, $Bbox2Caption$ task maintains a slight improvement.

\textbf{Freezing the vision encoder in the adaptation stage.} We freeze the vision encoder and only tune the LLM for adaptation and transfer in stage 2, enabling the LLM to better use the fine-grained vision feature for the $Caption2Bbox$ task. The vision encoder is kept frozen because it has already been sufficiently pretrained in stage 1, and further tuning it would significantly increase adaptation and transfer costs but yield marginal performance gains, as shown in Table~\ref{table-a5} in the appendix.

% As indicated in Table~\ref{table-a5} in the appendix, tuning vision encoder in stage 2 does not yield a significant performance improvement. This is because the vision encoder is trained well for fine-grained feature extraction in stage 1 and the $Caption2Bbox$ task relies primarily on the localization capability of the LLM. Therefore, to reduce the complexity of transfer and adaptation in stage 2, we keep the vision encoder frozen.

\subsection{Self Distillation for Fine-Grained Pretraining Paradigm}
Although the vision encoder is implicitly supervised through the caption loss $L_{caption}$ from the output text of LLM, it lacks explicit constraints on localized region features. Therefore, we incorporate self-distillation training \citep{naeem2024silc,maninis2024tips,zhang2019your} into stage 1. As illustrated in Fig.~\ref{fig:3}, an additional frozen teacher vision encoder is introduced to constrain the feature generated by the student vision encoder, thereby enhancing localized region features.

Specifically, given images $x$, the student vision encoder extracts the fine-grained features $x'$. Both prompts and visual features $x'$ are sent to the LLM for generating localized captions. Besides, the image regions $x_{crop}$ are cropped from $x$ according to the bbox coordinates and are sent to the teacher vision encoder for localized feature extraction. The self-distillation loss is calculated between $x'$ and $x'_{crop}$ by $L_{distill}=MSE(x'_{crop},ROIAlign(x'))$, where $x'_{crop}$ and $MSE$ denotes the extracted features of $x_{crop}$ and mean square error loss, respectively. The weights of the teacher vision encoder are initialized from the student vision encoder and updated by the exponential moving average (EMA) according to the student vision encoder, specifically $\theta_{tea}=\alpha \theta_{tea}+(1-\alpha)\theta_{stu}$. The overall loss can be written as $L=L_{caption}+\lambda L_{distill}$, where $\lambda$ denotes the weighting coefficient.

\section{Experiments}
\subsection{Experimental Settings}
\textbf{Dataset.}
Three datasets are adopted for distinct purposes of projector pre-training, fine-grained pretraining, and downstream SFT. Following \citep{liu2023visual}, we adopt the BLIP-LAION-CC-SBU-558K dataset \citep{huggingfaceLiuhaotianLLaVAPretrainDatasets} for projector pretraining. For fine-grained pretraining, we employ the constructed \emph{Gran-29M} dataset. To balance natural and OCR images, we proportionally sample 50M local region captions from OCR regions and use all the natural image local captions. During stage 1, we use all global samples and approximately 130M $Bbox2Caption$ samples. In stage 2, we sample 24M $Caption2Bbox$ samples in addition to global samples to reduce the transfer overhead. For SFT, we adopt the Open-LLaVA-NeXT 1M dataset \citep{chen2024open} for downstream adaptation. 

\textbf{Implementation Details.}
We use the LLaVA-Next framework \citep{li2024llava} for pre-training and SFT of the vision encoder and LLM. By default, we initialize the vision encoder with SigLIP2 \citep{tschannen2025siglip}, and adopt Qwen2.5-VL-1.5B \citep{bai2025qwen2} as the LLM. The projector is implemented with a two-layer MLP. In training, images are resized to 512$\times$512, padded according to their aspect ratio, and then fed into the vision encoder and LLM for feature extraction and inference. We also implement GranViT with image tiling strategy, as shown in Table~\ref{table-a2} in the appendix. We employ AdamW optimizer \citep{loshchilov2017decoupled} for pretraining and SFT with learning rate $10^{-5}$ for one epoch. The overall batch size is set to 256 with 128 Ascend 910B NPUs. $\lambda$ is 1 and $\alpha$ is 0.9 by default in our experiment. Abation study on $\lambda$ and $\alpha$ is shown in Table~\ref{table-a6} in the appendix.

\begin{table}[!t]
\renewcommand{\baselinestretch}{0.8}
\setlength{\abovecaptionskip}{0pt}
\setlength{\tabcolsep}{2.5pt}
\centering
\caption{Performance comparison with low resolution version. The bold font represents the best performance, and the underline represents the second performance.}\label{table2}
\begin{tabular}{c|l|ccccccc}
    \hline
     Capability&Benchmark &CLIP & SigLip & SigLip2 & AIMv2 &InternViT & SAILViT & GranViT \\
    \hline
    \multirow{11}{*}{Fine-Grained} & RefCOCO\_{testA} & $81.26$ & $69.47$ & $87.78$ & $86.03$ & $85.15$ & \underline{89.65} & \textbf{91.79} \\
                                               & RefCOCO\_{testB} & $64.51$ & $56.78$ & $76.90$ & $73.54$ & $71.40$ & \underline{79.82} & \textbf{83.88} \\
                                               & RefCOCO\_{val} & $74.71$ & $63.71$ & $83.26$ & $80.28$ & $78.48$ & \underline{85.32} & \textbf{89.13} \\
                                               & RefCOCO+\_{testA} & $74.43$ & $63.13$ & $82.92$ & $81.33$ & $77.33$ &\underline{85.01}& \textbf{87.04} \\
                                               & RefCOCO+\_{testB} & $51.25$ &$44.65$ & $66.47$ & $62.50$ & $58.58$ & \underline{69.66} & \textbf{73.24} \\
                                               & RefCOCO+\_{val} & $65.25$ & $55.84$ & $75.46$ &  $72.84$ & $69.18$ & \underline{78.10} & \textbf{81.55} \\
                                               & RefCOCOg\_{val} & $68.95$ & $55.02$ & $78.53$ & $76.55$ & $71.62$ & \underline{80.26} &\textbf{83.86} \\
                                               & RefCOCOg\_{test} & $68.52$ &$55.57$ &$78.94$ & $75.78$ & $72.71$ & \underline{80.92} & \textbf{83.82} \\
                                               & BLINK* & $51.87$ & $50.67$ & $50.35$ & $52.59$ & \underline{52.62} & $52.54$ &\textbf{56.80} \\
                                               & MMVP & $63.33$&$61.66$ & $66.00$ & $65.66$ & $61.00$ & \textbf{69.00} &\underline{66.33} \\
    \rowcolor{mygray}                                           & Average & $66.41$ & $57.67$ & $75.61$ & $73.50$ & $70.53$ & \underline{77.95} & \textbf{80.78} \\
    \hline
   \multirow{6}{*}{VQA} & MMBench & $61.14$ & $55.95$ & $64.00$ & \textbf{65.40} & $59.44$ & \underline{63.54}  &$62.46$ \\
                                   & MMStar & $39.46$ & $39.93$ & $39.33$ & $42.53$ & $40.20$ & \textbf{43.86} &\underline{43.73} \\
                                   & HallusionBench & $28.83$ & $28.22$ & \textbf{31.79} & $29.39$ & $27.95$ & \underline{31.29}  & $30.34$ \\
                                   & GQA& $58.80$ & $57.52$ & $60.42$ & $60.14$ & $58.02$ & \underline{60.83} &\textbf{60.95} \\
                                   & SEEDBench & $66.89$ & $66.28$ & $69.30$ & \underline{70.20} & $66.79$ & $69.75$ &\textbf{70.36} \\
    \rowcolor{mygray}                            & Average & $51.02$ & $49.58$ & $52.97$ & $53.53$ & $50.48$ & \textbf{53.85} &\underline{53.57} \\
    \hline
    \multirow{6}{*}{Reasoning} & MMMU & $40.00$ & $38.66$ & \underline{42.00} & $38.66$ & \textbf{44.00} & $38.66$ & $38.00$ \\
                                          & MathVista MINI & $38.10$ & $35.50$ & $38.40$ & $37.70$ & $39.50$ & \textbf{41.70} &\underline{40.40} \\
                                          & MMVet & $33.53$ & $30.87$ & \textbf{38.80} & \underline{38.34} & $35.27$ & $35.73$ & $37.29$ \\
                                          & ScienceQA & \underline{67.87} & $66.53$ & $66.98$ & $67.57$ & $66.63$ & \textbf{72.78}  &$67.42$ \\
                                          & AI2D & $66.51$ & $65.47$ & $69.26$ & $68.19$ & $66.51$ & \textbf{71.21}  &\underline{69.91} \\
    \rowcolor{mygray}                                      & Average & $49.20$ & $47.41$ & \underline{51.09} & $50.09$ & $50.38$ & \textbf{52.02} &$50.60$ \\
    \hline
    \multirow{6}{*}{OCR} & OCRBench & $406$ & $365$ & $515$ & $498$ & $461$ & \textbf{590} &\underline{551} \\
                         & DocVQA & $35.34$ & $34.26$ & $56.32$ & $50.56$ & $44.95$ & \underline{58.75} & \textbf{67.92} \\
                         & ChartQA & $50.84$ & $51.96$ & $64.44$ & $61.36$ & $60.68$ & \underline{63.24} & \textbf{67.96} \\
                         & InfoVQA & $20.67$ & $20.89$ & $24.12$ & $22.60$ & $21.91$ & \underline{24.75} &\textbf{27.19} \\
                         & TextVQA & $46.65$ & $41.71$ & \underline{61.36} & $56.57$ & $51.74$ & $60.90$ &\textbf{61.66} \\
    \rowcolor{mygray}                     & Average & $38.82$ & $37.06$ & $51.55$ & $48.18$ & $45.08$ & \underline{53.33} & \textbf{55.97} \\
    \hline

    \end{tabular}
    \vspace{-0.33cm}
    \end{table}

\subsection{Benchmark Evaluation}
We make extensive evaluations on the well-known OpenCompass benchmark \citep{contributors2023opencompass} and additional fine-grained benchmarks. We focus on fine-grained and OCR benchmarks that are divided into four classes: fine-grained (RefCOCO \citep{yu2016modeling}, RefCOCO+ \citep{yu2016modeling}, RefCOCOg \citep{yu2016modeling}, BLINK* \citep{fu2024blink}\footnote{We calculate the average score of fine-grained evaluation (Counting, Object Localization and Spatial Relation) in BLINK, denoted as BLINK*} and MMVP \citep{tong2024eyes}), mulitmodal VQA (MMBench \citep{liu2024mmbench}, MMStar \citep{chen2024we}, HallusionBench \citep{guan2024hallusionbench}, GQA \citep{hudson2019gqa} and SEEDBench \citep{li2023seed}), multimodal reasoning (MMMU \citep{yue2024mmmu}, MathVista MINI \citep{lu2023mathvista}, MMVet \citep{yu2023mm}, ScienceQA \citep{lu2022learn} and AI2D \citep{kembhavi2016diagram}) and OCR understanding (OCRBench \citep{liu2024ocrbench}, DocVQA \citep{mathew2021docvqa}, ChartQA \citep{masry2022chartqa}, InfoVQA \citep{mathew2022infographicvqa} and TextVQA \citep{singh2019towards}). We compare GranViT with diverse vision encoders, \emph{i.e.}, CLIP \citep{radford2021learning}, SigLip \citep{zhai2023sigmoid}, SigLip2 \citep{tschannen2025siglip}, AIMv2 \citep{fini2025multimodal}, InternViT \citep{chen2024internvl}, and SAILViT \citep{yin2025sailvit}.

\textbf{Performance Comparison.}
Table~\ref{table2} provides a performance comparison on various benchmarks. For fine-grained and OCR tasks, GranViT achieves an average top-1 score of 80.78 and 55.97, and surpasses the second best by 2.83 and 2.64, respectively. GranViT is comparable to SAILViT with a marginal difference of only 0.3 in multimodal VQA tasks. For multimodal reasoning tasks, GranViT suffers a slight performance loss of 0.4 compared to SigLIP2. This is because reasoning capability does not rely heavily on fine-grained feature extraction, and GranViT prioritizes fine-grained tasks rather than extensively training reasoning capabilities. Note that reasoning capability can be further improved by applying reasoning VQA data in pretraining.

\begin{table}[!t]
\renewcommand{\baselinestretch}{0.5}
\setlength{\tabcolsep}{2.5pt}
\setlength{\abovecaptionskip}{0pt}
\centering
\caption{Performance comparison for transferring vision encoders to Qwen2.5-3B and Qwen2.5-7B. The best results are highlighted in bold and the second best underlined. Ref, Ref+ and Refg denote the RefCOCO\_{testA}, RefCOCO+\_{testA} and RefCOCOg\_{test}. MMB, HB, and SB stand for MMBench, HallusionBench, and SEEDBench, and. SQA, OB, DVQA, and IVQA for ScienceQA, OCRBench, DocVQA, and InfoVQA, respectively.}\label{table3}
\begin{tabular}{l|cccccccccccc}
    \hline
    Model & Ref & Ref+ & Refg  & MMB & HB & SB & MMMU  & SQA & OB & DVQA & IVQA & Avg  \\
    \hline                    
    \rowcolor{mygray}\multicolumn{13}{c}{Qwen2.5-3B} \\
    \hline
    CLIP & $86.60$ & $81.73$ & $75.26$ & $65.09$ & $31.67$ & $68.55$ & \underline{39.33} & $73.12$ & $413$& $38.70$ & $24.08$ & 56.86 \\
    SigLip & $87.55$ & $82.83$& $77.46$ & $69.27$ & $33.12$ & $70.29$ & $36.44$  & $70.64$ & $428$ & $46.27$ & $25.24$ &58.36 \\
    SigLip2 & $91.03$& $86.37$  & $83.30$ & \underline{69.34} & $33.70$ & $71.13$ & $36.00$ & $71.69$ & $529$ & $60.87$ & $29.00$ &62.30 \\
    AIMv2 & $90.33$ & $87.05$ & $81.83$ & $69.27$ & $31.91$ & $71.84$ & \textbf{40.00} & \underline{74.46} & $545$ & $56.04$ & $27.34$ &62.23 \\
    SAILViT & \underline{91.58} & \underline{87.82} & \underline{83.63} & \textbf{69.73} & \textbf{34.08} & $71.33$ & $36.00$  & \textbf{75.75}& \textbf{633} & \underline{62.53} & \underline{29.49} &\underline{64.11} \\
    GranViT & \textbf{93.22} & \textbf{89.32} & \textbf{86.17} & $67.56$ & \underline{33.77} & \textbf{72.34} & $38.66$ & $73.24$ & \underline{590} & \textbf{71.09} & \textbf{29.96} &\textbf{64.94} \\
    \hline
    \rowcolor{mygray}\multicolumn{13}{c}{Qwen2.5-7B} \\
    \hline
    CLIP & $90.01$ & $86.25$ & $80.44$ & $70.58$ & $36.39$ & $70.70$ & \textbf{46.00} & $75.26$ & $466$ & $43.02$ & $24.88$ & 60.92\\
    SigLip & $90.68$ & $86.27$ & $81.72$ & \underline{74.22} & \underline{37.94} & $72.04$ & \textbf{46.00} & \underline{76.99} & $459$ & $50.03$ & $26.69$ &62.59 \\
    SigLip2 & $92.06$ & $88.70$ & $85.13$ & $72.21$ & $36.02$& $72.17$ & $44.00$ & $75.26$ & $540$ & $62.20$ & $29.89$ &64.69\\
    AIMv2 & $90.84$ & $87.82$ & $82.39$ & $72.29$ & $37.29$ & $72.11$ & $41.44$ & $72.92$ & $553$ & $56.54$ & $28.59$ &63.41\\
    SAILViT & \underline{92.66} & \underline{89.50} & \underline{85.32} & \textbf{74.53} & $37.76$ & \underline{73.05} & \underline{44.66} & \textbf{81.11} & \textbf{648} & \underline{64.13} & \underline{30.66} & \underline{67.11} \\
    GranViT & \textbf{92.98} & \textbf{90.46} & \textbf{87.96} & $73.37$ & \textbf{39.37} & \textbf{74.45} & \underline{44.66} &  $75.85$ & \underline{582} & \textbf{73.14} & \textbf{31.69} & \textbf{67.47}\\
    \hline
\end{tabular}
\vspace{-0.33cm}
\end{table}
    
\textbf{Transferability.} Table~\ref{table3} reports the performance comparison with larger LLMs (\emph{i.e.}, Qwen2.5-3B, Qwen2.5-7B). Notably,  other vision encoders directly employ the larger LLM for SFT, whereas GranViT employs a lightweight LLM (\emph{i.e.}, Qwen2.5-1.5B) for pre-training in stage 1, transfers to the larger LLM in stage 2, and subsequently undergoes SFT. GranViT also demonstrates outstanding performance on fine-grained and OCR tasks, while achieving comparable or even state-of-the-art performance on some VQA tasks (\emph{i.e.}, HallusionBench and SEEDBench).

\subsection{Ablation Study}
We employ small-scale datasets for ablation studies on the contribution of each module. 8 million global and $Bbox2Caption$ QA pairs and 8 million global and $Caption2Bbox$ QA pairs are sampled from \emph{Gran-29M} as training data for two stages, respectively. The entire dataset is used for training the projector and conducting supervised fine-tuning (SFT).

\begin{table}[!t]
\renewcommand{\baselinestretch}{0.8}
\renewcommand{\arraystretch}{0.8}
\setlength{\abovecaptionskip}{0pt}
\setlength{\tabcolsep}{5pt}
\centering
\caption{Ablation study on each component of the proposed GranViT.}\label{table4}
\begin{tabular}{cccc|c|c|c|c}
    \hline
     SigLip2 & Stage1 & Self-Distillation & Stage2 & Fine-Grained & VQA & Reasoning & OCR \\
    \hline
    \Checkmark&\XSolidBrush&\XSolidBrush&\XSolidBrush& $73.20$ & $52.97$ & \textbf{51.09} & $51.55$ \\
    \Checkmark&\Checkmark&\XSolidBrush&\XSolidBrush&$75.06$ & $53.64$ & $49.89$ & $52.77$ \\
    \Checkmark&\Checkmark&\Checkmark&\XSolidBrush&\underline{75.55}&\textbf{53.90} & \underline{50.32} & \underline{53.02}\\
    \Checkmark&\Checkmark&\Checkmark&\Checkmark&\textbf{76.54} & \underline{53.77} & $48.99$ & \textbf{53.78}\\
    \hline
    \end{tabular}
    \vspace{-0.33cm}
    \end{table}

\textbf{Effectiveness of Training Paradigm.} In Table~\ref{table4}, the progressive introduction of the two-stage training strategy along with self-distillation results in incremental performance gains on fine-grained and OCR-related tasks. Specifically, with stage 1 pretraining, the MLLM exhibits a substantial improvement in both fine-grained recognition capability and OCR understanding (2.2 and 1.2 gains). Self-distillation training further improves fine-grained and OCR evaluations. With stage 2 adaptation, fine-grained and OCR evaluations yield additional gains of 1.0 and 0.7, respectively.
    
\begin{table}[!t]
\renewcommand{\baselinestretch}{0.8}
\renewcommand{\arraystretch}{1.0}
\setlength{\abovecaptionskip}{0pt}
\setlength{\tabcolsep}{8pt}
\centering
\caption{Performance with different vision encoder initialization for GranViT during pretraining.}\label{table6}
\begin{tabular}{l|c|c|c|c}
    \hline
     Model & Fine-Grained & VQA & Reasoning & OCR \\
    \hline
    InternViT & $69.76$ & $50.48$ & \textbf{50.38} & $45.08$ \\
    GranViT (InternViT) & \textbf{75.15} & \textbf{51.78} & 50.23 & \textbf{50.13} \\
    \hline
    AIMv2 & $72.28$ & $53.53$ & $50.09$ & $48.18$ \\
    GranViT (AIMv2) & \textbf{77.14} & \textbf{55.07} & \textbf{50.59} & \textbf{52.71} \\
    \hline
    SAILViT & $75.42$ & $53.85$ & \textbf{52.02} & $54.53$ \\
    GranViT (SAILViT) & \textbf{76.79} & \textbf{55.40} & 51.95 & \textbf{56.61} \\
     \hline 
    \end{tabular}
    \vspace{-0.33cm}
    \end{table}
    
\textbf{Different Initialization for Vision Encoder.} In Table~\ref{table6}, we compare the performance of vision encoders initialized with different models. All three distinct vision encoders (InternViT-300M \citep{chen2024internvl}, AIMv2 \citep{fini2025multimodal}, and SAILViT-Huge \citep{yin2025sailvit}) exhibit significant performance improvements after pre-training, with the most notable improvements in fine-grained perception (\emph{i.e.}, 5.3 for InternViT, 4.8 for AIMv2, and 1.3 for SAILViT) and OCR understanding (\emph{i.e.}, 5.1 for InternViT, 4.5 for AIMv2, and 2.1 for SAILViT). 

\textbf{Visualization.} To illustrate the fine-grained feature extraction capability of GranViT, we visualize in Fig.~\ref{fig:1}(b) the attention maps of different vision encoders. AIMv2 \citep{fini2025multimodal} and SAILlViT \citep{yin2025sailvit} focus on global features and are severely deficient in local regions. SigLip2 \citep{tschannen2025siglip} emphasizes local regions, but exhibits redundant attention to global features. In contrast, GranViT can simultaneously consider local regions and exclude interference from redundant features. This further validates the effectiveness of the proposed pre-training framework.

\section{Conclusion}
This paper proposed GranViT, a novel visual Transformer architecture that integrates fine-grained perception and multimodal alignment for advanced multimodal understanding. GranViT is trained on \emph{Gran-29M}, a newly curated large-scale dataset containing global and region-level descriptive annotations for both natural and OCR images. The region-level bounding box and text annotations enable two dedicated tasks, \emph{i.e.}, $Bbox2Caption$ for optimizing the vision encoder to strengthen fine-grained feature extraction and $Caption2Bbox$ for adapting vision features to different LLMs with enhanced region localization. Self-distillation loss is further incorporated to explicitly enhance local feature learning. GranViT is potential to serve as a robust foundation MLLM model that offers strong capabilities for complex multimodal reasoning tasks.

\bibliography{iclr2026_conference}
\bibliographystyle{iclr2026_conference}

\appendix
\section{Appendix}

\subsection{Details about \emph{Gran-29M}}

\begin{table}[!t]
\renewcommand{\baselinestretch}{1.0}
\renewcommand{\arraystretch}{1.0}
\setlength{\tabcolsep}{12.0pt}
\setlength{\abovecaptionskip}{0pt}
\centering
\caption{Detailed data sources of datasets used in \emph{Gran-29M}. }\label{table-a1}
\begin{tabular}{c l r r}
\hline
Data Type & Data Source & $\#images$ & $\#regions$ \\
\hline
\multirow{8}{*}{Natural} &  CC3M & 565521 & 2342622 \\
                         & IN21k & 614367 & 1628363 \\
                         &  LAION & 17194230 & 54356988 \\
                         & SBU & 21479 & 52259 \\
                         & CC12M & 4909682 & 21714139 \\
                         & FLICKR30k & 1269 & 4351 \\
                         & YFCC15M & 655400 & 1884172 \\
                         & VisualGenome & 2150 & 14825 \\
\hline
\multirow{30}{*}{OCR}  & Arxiv & 2655630 & 22574019 \\
                      & InfoVQA & 4343 & 21129 \\
                      & LRV-Instruction & 8304 & 22152 \\
                      & OCRVQA & 86 & 238 \\
                      & PDFVQA & 12253 & 98783 \\
                      & POIE & 898 & 12881 \\
                      & SROIE & 994 & 25586 \\
                      & PubTables\_en & 121330 & 522516 \\
                      & RenderedText & 7031 & 20003 \\
                      & MMC-Instruction & 58583 & 218288 \\
                      & AI2D\_gpt4v & 1958 & 27919 \\
                      & AI2D\_internvl & 11981 & 110438 \\
                      & ArxivQA & 52517 & 1532997 \\
                      & Chart2Text & 22051 & 698874 \\
                      & Diagram\_Image\_To\_Text & 154 & 2244 \\
                      & Robut\_SQA & 5714 & 740443 \\
                      & Robut\_WikiSQL & 38935 & 7092396 \\
                      & Docx\_en & 429182 & 3720371 \\
                      & AI2D\_original & 2364 & 27348 \\
                      & FigureQA & 96000 & 1316177 \\
                      & Hitab & 2495 & 392662 \\
                      & Robut\_wtq & 38241 & 5677381 \\
                      & TextCaps & 20548 & 186867 \\
                      & TextOCR & 23511 & 215809 \\
                      & Uber\_Text & 118042 & 571779 \\
                      & CORD & 955 & 3482 \\
                      & ChartQA & 14650 & 45871 \\
                      & DocBank & 25482 & 172933 \\
                      & SynthText & 756552 & 3563689 \\
                      & Docmatrix & 1019766 & 51945528 \\
\hline
  \end{tabular}%}
\end{table}

\begin{table}[!t]
\renewcommand{\baselinestretch}{1.0}
\renewcommand{\arraystretch}{1.2}
\setlength{\tabcolsep}{8.0pt}
\setlength{\abovecaptionskip}{15pt}
\centering
\caption{Data sources of natural and OCR images in \emph{Gran-29M}. $\#images$ and $\#regions$ denote the number of images and annotated bounding boxes after filtering, respectively.}\label{table-a7}
\begin{tabular}{c l r r}
\hline
Data Type & Data Source & $\#images$ & $\#regions$ \\
\hline
% \multirow{8}{*}{Natural} & CC3M  & 565K & 2.3M \\
%                          & IN21k & 614K & 1.6M \\
%                          & LAION & 17.19M & 54.35M \\
%                          & SBU & 21K & 52K \\
%                          & CC12M  & 4.9M & 21.74M \\
%                          & FLICKR30k  & 1269 & 4351 \\
%                          & YFCC15M & 655K & 1.8M \\
%                          & VisualGenome  & 2150 & 14825 \\

\multirow{3}{*}{Natural} & UMG-41M  & 6.7M & 27.63M \\
                         & LAION & 17.19M & 54.35M \\
                         & FLICKR30k  & 1269 & 4351 \\
                         
\hline
\multirow{4}{*}{OCR}  & Text Images & 1.9M & 42.57M \\
                      & Chart, Table & 325K & 2.8M \\
                      & Invoice Receipt & 2847 & 41K \\
                      & Rich Text Images & 3.3M & 56M \\
\hline
& TOTAL & 29.51M & 183.55M \\
\hline
\end{tabular}
\vspace{-0.5cm}
\end{table}

We systematically document the data sources of both natural and OCR images utilized in the \emph{Gran-29M} dataset, as shown in Table~\ref{table-a1} and Table~\ref{table-a7}. For natural images, CC3M \citep{sharma2018conceptual}, IN21k \citep{deng2009imagenet}, SBU \citep{ordonez2011im2text}, CC12M \citep{changpinyo2021conceptual}, YFCC15M \citep{kamath2021mdetr}, VisualGenome \citep{krishna2017visual}, together with LAION \citep{schuhmann2022laion} and FLICKR30k \citep{young2014image}, are contained. For OCR images, there are 30 datasets contained in toal for diversity: Arxiv, InfoVQA \citep{mathew2022infographicvqa}, LRV-Instruction \citep{liu2023aligning}, OCRVQA \citep{mishra2019ocr}, PDFVQA \citep{ding2023vqa}, POIE \citep{kuang2023visual}, SROIE \citep{huang2019icdar2019}, PubTables\_en \citep{smock2022pubtables}, RenderedText \citep{huggingfaceWendlercRenderedTextDatasets}, MMC-Instruction \citep{liu2023mmc}, AI2D\_gpt4v \citep{huggingfaceAbhayzalaAI2DCaptionDatasets}, AI2D\_internvl \citep{chen2024internvl}, ArxivQA \citep{li2024multimodal}, Chart2Text \citep{kantharaj2022chart}, Diagram\_Image\_To\_Text, Robut\_SQA \citep{zhao2023robut}, Robut\_WikiSQL \citep{zhao2023robut}, Docx\_en, AI2D\_original \citep{kembhavi2016diagram}, FigureQA \citep{kahou2017figureqa}, Hitab \citep{cheng2021hitab}, Robut\_wtq \citep{zhao2023robut}, TextCaps \citep{sidorov2020textcaps}, TextOCR, Uber\_Text, CORD \citep{park2019cord}, ChartQA \citep{masry2022chartqa}, DocBank \citep{li2020docbank}, SynthText \citep{gupta2016synthetic} and Docmatrix \citep{laurenccon2024building}.

\subsection{High resolution performance of GranViT}
Additionally, we provide the evaluation results of GranViT with an image tiling strategy in the pretraining. According to image tiling\citep{chen2024internvl,gu2024infinity,wang2024qwen2,lu2024deepseek,team2025kwai}, images are firstly converted into N $512\times512$ local patches and one global patch. All patches are simultaneously fed into the vision encoder for feature extraction. With a patch size of $16$, we obtain $N\times1024$ visual patch features. We use pixel shuffle \citep{gu2024infinity,wang2024qwen2}to compress these visual features to $N\times256$ patches. These visual features are then recombined based on positions and fed into the projector and LLM for understanding. The evaluation results are reported in Table~\ref{table-a2}. 

\begin{figure}[!t]
    \centering
\includegraphics[height=7cm,width=13cm]{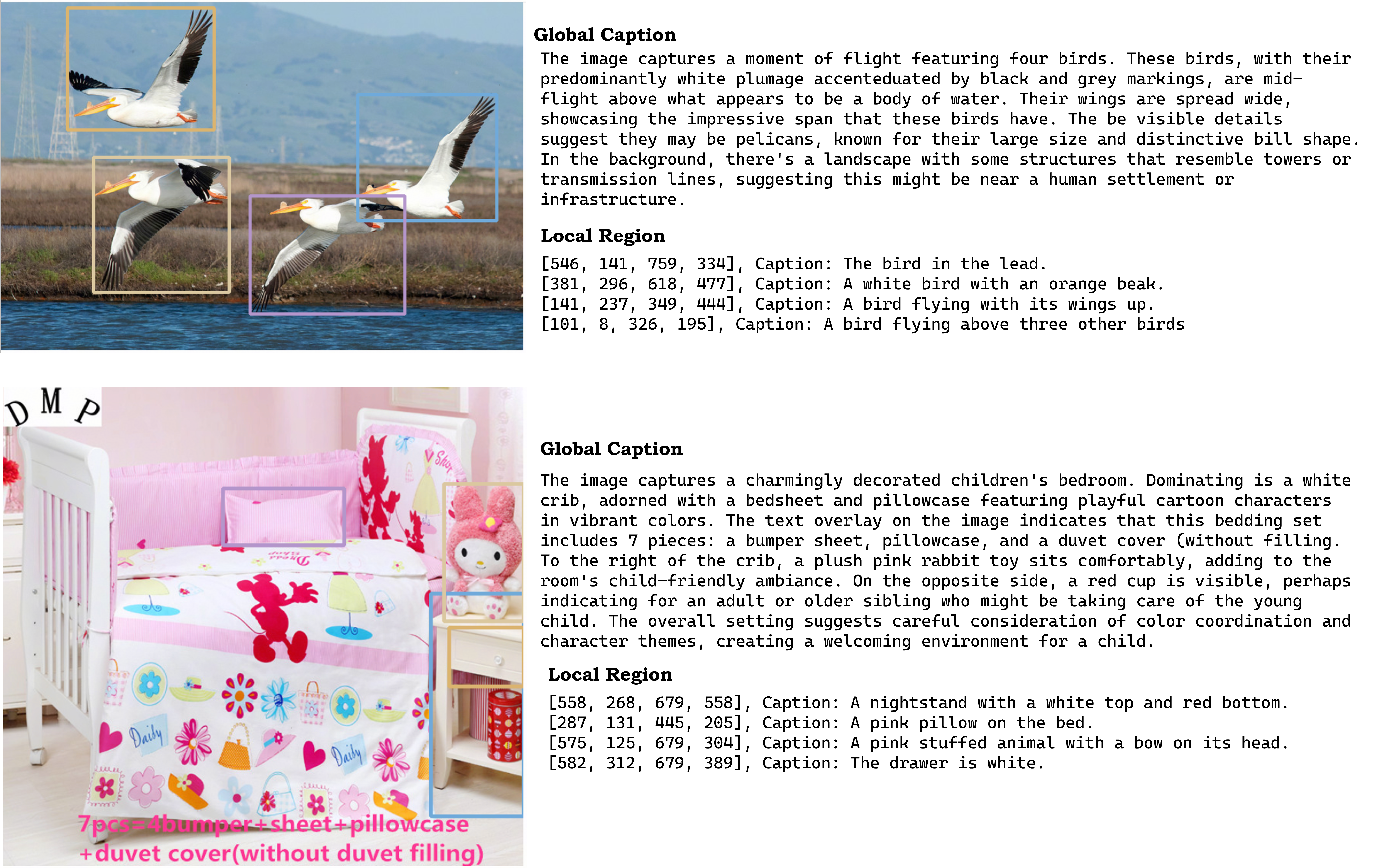}
    \caption{Visualization of \emph{Gran-29M}.}
    \label{fig:A-5}
\end{figure}

\begin{figure}[!t]
    \centering
\includegraphics[height=7cm,width=13cm]{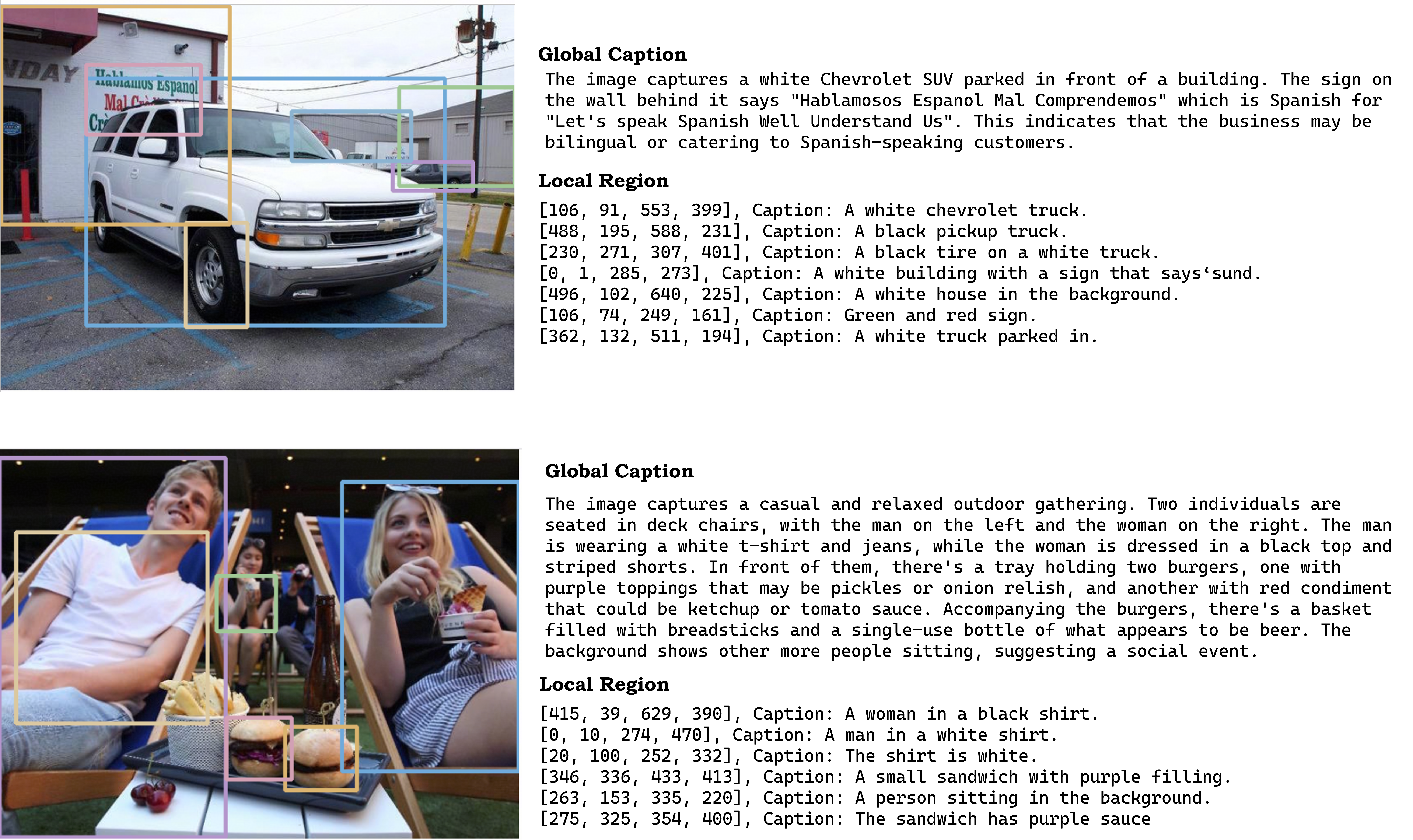}
    \caption{Visualization of \emph{Gran-29M}.}
    \label{fig:A-6}
\end{figure}

\begin{table}[!t]
\renewcommand{\baselinestretch}{0.8}
\renewcommand{\arraystretch}{1.3}
\setlength{\tabcolsep}{3pt}
\centering
\caption{Performance comparison with image tiling. The bold font represents the best performance, and the underline represents the second performance.} \label{table-a2}
\begin{tabular}{c|l|cccccc}
    \hline
     Capability&Benchmark &CLIP & SigLip & SigLip2 & AIMv2 & SAILViT & GranViT \\
    \hline
    \multirow{11}{*}{Fine-Grained} & RefCOCO\_{testA} & $82.03$ & $82.58$ & $84.95$ & $84.58$ & \underline{87.92} & \textbf{90.71}\\
                                   & RefCOCO\_{testB} & $66.53$ & $67.47$ & $74.91$ & $70.02$ & \underline{76.85} & \textbf{82.04} \\
                                   & RefCOCO\_{val} & $76.01$ & $77.25$ & $80.11$ & $77.97$ & \underline{83.12} & \textbf{87.21} \\
                                   & RefCOCO+\_{testA} &$75.16$ & $77.68$ & $80.24$ & $79.23$ & \underline{82.48} & \textbf{85.15} \\
                                   & RefCOCO+\_{testB} & $32.00$ & $55.67$ & $63.81$ & $59.97$ & \underline{67.13} & \textbf{70.52} \\
                                   & RefCOCO+\_{val} & $65.74$ & $68.32$ & $72.26$ & $70.71$ & \underline{75.59} & \textbf{79.05}\\
                                   & RefCOCOg\_{val} &  $70.73$ & $72.74$ & $75.23$ & $74.91$ & \underline{78.92} &\textbf{81.98} \\
                                   & RefCOCOg\_{test} &  $70.81$ & $72.18$ & $74.75$ & $74.48$ & \underline{77.98} &\textbf{81.63} \\
                                   & BLINK* & $51.40$ & $52.84$ & \underline{53.49} & \textbf{56.25} & $52.33$ &52.42 \\
                                   & MMVP & $61.33$ & $64.00$ & $65.33$ & \underline{67.33} & \textbf{68.00}  &65.66\\
    \rowcolor{mygray}                               & Average & $67.31$ & $69.07$ & $72.51$ & $71.55$ & \underline{75.03} &\textbf{77.64} \\
    \hline
   \multirow{6}{*}{VQA} & MMBench & $60.44$ & $64.39$ & \underline{64.62} & \textbf{65.01} & $64.00$ & 63.77 \\
                                   & MMStar & $39.06$ & $41.26$ & \underline{42.73} & $41.80$ & \textbf{45.60} &40.80 \\
                                   & HallusionBench & $30.18$ & $29.89$ & $30.08$ & $26.95$ & \underline{30.56} &\textbf{35.93} \\
                                   & GQA& $58.99$ & $59.88$ & $59.95$ & $60.04$ & \underline{60.72} & \textbf{61.32} \\
                                   & SEEDBench &  $67.35$ & $68.75$ & \underline{69.60} & $69.52$ & \textbf{70.07} &69.54 \\
    \rowcolor{mygray}                               & Average & $51.20$ & $52.83$ & $53.40$ & $52.66$ & \underline{54.19} &\textbf{54.27} \\
    \hline
    \multirow{6}{*}{Reasoning} & MMMU & $38.11$ & $35.44$ & \textbf{41.33} & \underline{40.66} & $40.44$ &35.66 \\
                                          & MathVista MINI & $36.80$ & $35.80$ & $37.40$ & \underline{38.90} & $38.50$ &\textbf{40.00} \\
                                          & MMVet & $35.27$ & $32.47$ & \textbf{40.59} & \textbf{40.59} & \underline{40.13} & 38.30\\
                                          & ScienceQA &  $66.18$ & $68.66$ & $68.36$ & \underline{66.98} & \textbf{74.71} &66.28 \\
                                          & AI2D & $67.13$ & $68.65$ & $69.62$ & $69.52$ & \textbf{71.85} &\underline{69.88} \\
    \rowcolor{mygray}                                      & Average & $48.70$ & $48.20$ & \underline{51.46} & $51.33$ & \textbf{53.13} &50.02 \\
    \hline
    \multirow{6}{*}{OCR} & OCRBench & $414$ & $450$ & $545$ & $522$ & \underline{551} &\textbf{583} \\
                         & DocVQA & $53.16$ & $58.36$ & $68.74$ & $64.58$ & \underline{71.81} &\textbf{72.81} \\
                         & ChartQA & $60.12$ & $62.60$ & $65.04$ & $65.48$ & \underline{67.36} & \textbf{71.96} \\
                         & InfoVQA & $24.17$ & $27.81$ & $31.71$ & $31.38$ & \underline{33.47} &\textbf{33.59} \\
                         & TextVQA &  $56.74$ & $60.95$ & $67.78$ & $66.33$ & \textbf{69.47} &\underline{69.40} \\
    \rowcolor{mygray}                     & Average & $47.12$ & $50.94$ & $57.55$ & $55.99$ & \underline{59.44} & \textbf{61.21} \\
    \hline

    \end{tabular}
    \end{table}

\begin{figure}[!t]
    \centering
\includegraphics[height=7cm,width=13.5cm]{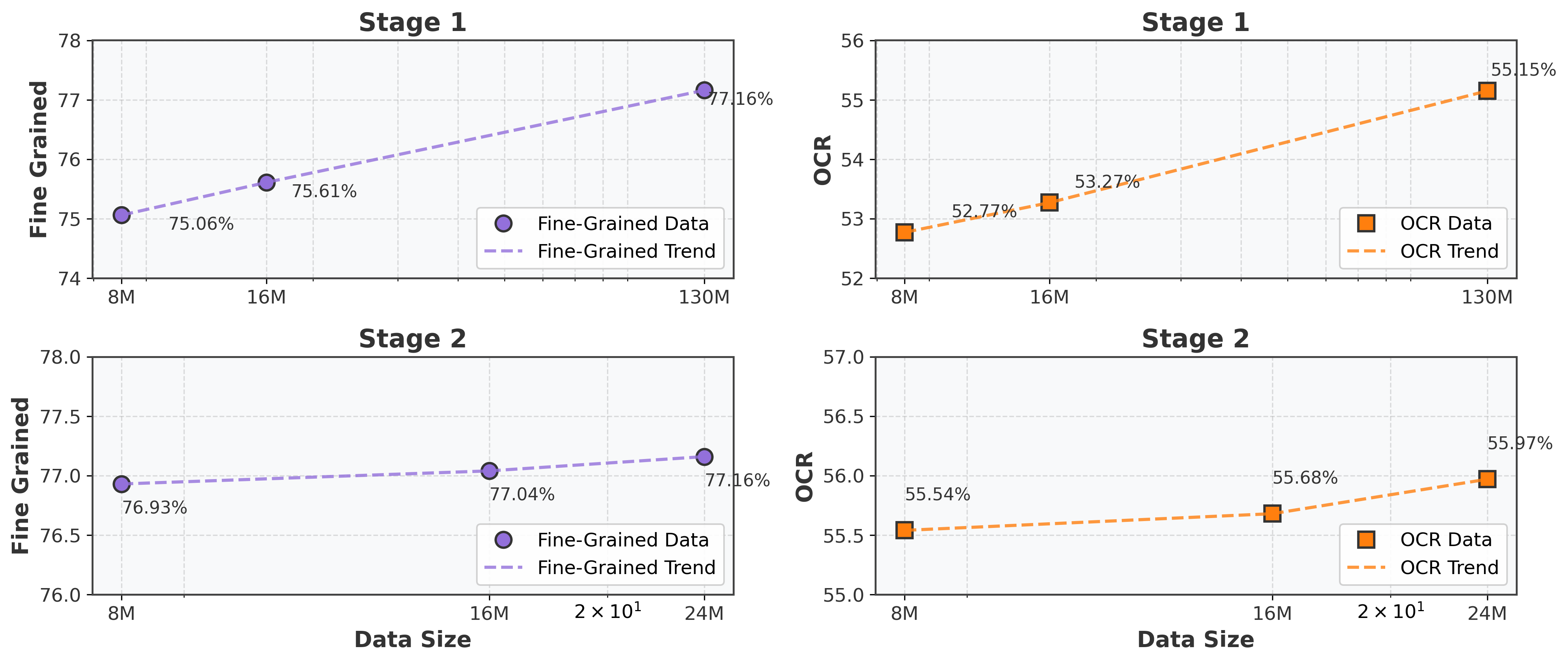}
\vspace{-0.3cm}
    \caption{Scaling law of two-stage training.}
    \label{fig:a1}
\end{figure}

\subsection{Scaling Laws} 
We evaluate the scaling capacity of pretraining-adaptation framework in Fig.~\ref{fig:a1}. Specifically, for stage 1, we leverage 8M, 16M and all the 130M regions for $Bbox2Caption$ tasks, while we leverage 8M, 16M, 24M and all the 130M regions for $Caption2Bbox$ tasks in stage 2. The average score of fine-grained (RefCOCO\_testA, RefCOCO+\_testA, RefCOCOg\_test, BLINK, MMVP) and OCR (OCRBench, DocVQA, ChartQA, InfoVQA, TextVQA) tasks is reported. As the data scale increases, both tasks exhibit significant performance improvements, indicating enhanced fine-grained feature extraction capability of GranViT.

\begin{table}[!t]
\renewcommand{\baselinestretch}{0.8}
\renewcommand{\arraystretch}{0.8}
\setlength{\tabcolsep}{3pt}
\centering
\caption{Performance comparison of whether the vision encoder is frozen in stage 2.}\label{table-a5}
\begin{tabular}{c|c|c|c|c}
    \hline
     Vision Encoder State & Fine-Grained & VQA & Reasoning & OCR \\
    \hline
    Frozen & $77.24$ & $54.83$ & $51.34$ & $54.02$ \\
    Tunable & $77.17$ & $54.83$ & $52.48$ & $54.06$ \\
    \hline
    \end{tabular}
    \vspace{-0.33cm}
    \end{table}
    
\subsection{The reason for freezing the vision encoder in stage 2}
In Table~\ref{table-a5}, we compare the performance gap between the vision encoder that is frozen and tunable in stage 2. Tuning the vision encoder in stage 2 does not achieve significant improvement, while leading to more training cost, since the vision encoder is trained well in stage 1 for fine-grained feature extraction. Therefore, to reduce the training complexity, we freeze the vision encoder in stage 2.

\subsection{The difference between GranViT and SAILViT} 
SAILViT \citep{yin2025sailvit} addresses the problem of insufficient visual-language alignment through a three-stage pre-training strategy that co-optimizes the vision encoder, projector, and LLM to achieve better alignment. GranViT differs from SAILViT in two key aspects. First, GranViT leverages both $Bbox2Caption$ and $Caption2Bbox$ tasks to strengthen fine-grained feature extraction and local region localization in the vision encoder and LLM, while SAILViT relies solely on global question-answering in the pretraining. Second, while SAILViT injects world knowledge into the vision encoder using large-scale SFT data to improve task-specific performance, GranViT focuses on enhancing the generic representation ability of the vision encoder. We contend that improved generic representations facilitate better adaptation to downstream tasks. Notably, the two strategies are complementary: after learning stronger generic features, task-specific adaptation following the paradigm of SAILViT can further enhance the performance of MLLM on specialized applications.

\begin{table}[!t]
\renewcommand{\baselinestretch}{0.8}
\renewcommand{\arraystretch}{0.8}
\setlength{\tabcolsep}{3pt}
\centering
\caption{Performance comparison when the vision encoder is frozen during SFT training.}\label{table-a3}
\begin{tabular}{c|c|c|c|c}
    \hline
     Vision Encoders & Fine-Grained & VQA & Reasoning & OCR \\
    \hline
    SigLip2 & $70.51$ & $53.36$ & $51.41$ & $49.16$ \\
    AIMv2 & $57.31$ & $52.24$ & $48.75$ & $46.90$ \\
    SAILViT & $71.90$ & $54.16$ & $50.93$ & $52.79$ \\
    GranViT & $75.16$ & $53.07$ & $49.12$ & $54.81$ \\
    \hline
    \end{tabular}
    \vspace{-0.33cm}
    \end{table}

\begin{table}[!t]
\renewcommand{\baselinestretch}{0.8}
\renewcommand{\arraystretch}{1.0}
\setlength{\tabcolsep}{2.5pt}
\centering
\caption{Continue training performance of GranViT. Ref, Ref+ and Refg denotes the RefCOCO\_{testA}, RefCOCO+\_{testA} and RefCOCOg\_{test} respectively. MMB, HB, and SB denote the MMBench, HallusionBench, and SEEDBench. SQA, OB, DVQA, and IVQA denote ScienceQA, OCRBench, DocVQA, and InfoVQA, respectively.}\label{table-a4}
\begin{tabular}{l|cccccccccccc}
    \hline
     Model & Ref & Ref+ & Refg  & MMB & HB & SB & MMMU  & SQA & OB & DVQA & IVQA & Avg  \\
    \hline
    SigLip2 & 88.43 & 83.26& 79.54&62.61&32.54&69.77 & 38.66 & 71.93& 584 & 60.56 & 26.44 & 61.10 \\ 
    \rowcolor{mygray} GranViT & \textbf{91.26}& \textbf{86.67} & \textbf{84.32} & 61.37 & 32.44&\textbf{70.48}  & 36.77& 70.59 & \textbf{623} & \textbf{68.16}& \textbf{28.61}& \textbf{62.99} \\
     \hline
    \end{tabular}
    \end{table}

\subsection{Continue Pretraining} 
We augment our two-stage training paradigm via continuous pretraining with SFT data, following stage 3 in SAILViT \citep{yin2025sailvit}. We leverage LLaVA-One-Vison \citep{li2024llavaonevision} dataset for pretraining and utilize Open-LLaVA-NeXT 1M dataset \citep{chen2024open} for SFT. GranViT is initialized from SigLip2 and pretrained with \emph{Gran-29M} dataset (8M samples for both stages) first. Then, we apply continuous pretraining with SFT data to both models. As shown in Table~\ref{table-a4}, GranViT outperforms SigLiP2 on fine-grained and OCR evaluation significantly. This experiment demonstrates that our method is compatible with the pretraining approach of SAILViT and that SFT data can be subsequently incorporated after our pretraining paradigm to further performance improvement.

\begin{table}[!t]
\renewcommand{\baselinestretch}{0.8}
\renewcommand{\arraystretch}{1.0}
\setlength{\tabcolsep}{5pt}
\centering
\caption{Ablation Study of the coefficient in self-distillation.}\label{table-a6}
\begin{tabular}{cc|c|c|c|c}
    \hline
     $\lambda$ & $\alpha$ & Fine-Grained & VQA & Reasoning & OCR \\
    \hline
    1 & 0.9 & $75.55$ & $53.90$ & $50.32$ & $53.02$ \\
    1 & 0.99 & $75.25$ & $53.45$ & $50.89$ & $53.30$\\
    1&0.999& $74.86$ & $53.53$ & $50.13$ & $52.31$ \\
    \hline
    0.01 & 0.9 & $74.81$ & $53.47$ & $50.13$ & $52.80$ \\
    0.1 & 0.9& $74.75$ & $53.48$ & $50.58$ & $53.16$ \\
    0.5& 0.9 & $75.02$ & $53.80$ & $51.65$ & $53.32$ \\
     \hline
    \end{tabular}
    \end{table}
    
\subsection{Coefficient in Self-Distillation} 
We ablate two parameters in the self-distillation process: $\lambda$ and $\alpha$. To efficiently conduct the ablation experiments, during the pre-training stage, we only train stage 1 and then directly proceed to downstream SFT. As shown in Table~\ref{table-a6}, the evaluation performance of the model on fine-grained tasks gradually improves as $\lambda$ increases and $\alpha$ decreases. Therefore, we set $\lambda$ to 1 and $\alpha$ to 0.9 in our experiments.
    
\subsection{Frozen Vision Encoder in SFT} 
To isolate the training gains of the vision encoder during the SFT stage, we compared the performance of different vision encoders with a frozen MLLM. As shown in Table~\ref{table-a3}, GranViT achieved the best performance in both fine-grained perception and OCR evaluations, outperforming SailViT by an average of $3.2$ and $2.1$, respectively. This fully demonstrates the significantly stronger fine-grained perception capability of GranViT.

\end{document}